\pdfoutput=1

\documentclass[11pt]{article}

\usepackage{etoolbox}
\newtoggle{arxiv}
\toggletrue{arxiv}

\iftoggle{arxiv}{
  \setlength{\textwidth}{6.5in}
  \setlength{\textheight}{9in}
  \setlength{\oddsidemargin}{0in}
  \setlength{\evensidemargin}{0in}
  \setlength{\topmargin}{-0.5in}
  \newlength{\defbaselineskip}
  \setlength{\defbaselineskip}{\baselineskip}
  \setlength{\marginparwidth}{0.8in}
}{
\usepackage[compact]{titlesec}
\titlespacing{\section}{0pt}{*1}{*0}
\titlespacing{\subsection}{0pt}{*1.5}{*0}

\usepackage[subtle, mathdisplays=tight, charwidths=normal, leading=normal]{savetrees}

\addtolength\textfloatsep{-0.5em}
\addtolength\intextsep{-0.2em}

\def\setstretch#1{\renewcommand{\baselinestretch}{#1}}
\setstretch{0.985}
\addtolength{\parskip}{-1pt}

}

\usepackage{enumitem}
\usepackage{times}
\usepackage{latexsym}
\usepackage{hyperref}
\usepackage{algorithm}
\usepackage{algorithmic}
\usepackage{amsmath}
\usepackage{amssymb}
\usepackage{mathtools}
\usepackage{amsthm}
\usepackage[T1]{fontenc}
\usepackage{subfigure}
\usepackage{tabu}
\usepackage{booktabs} 
\usepackage{colortbl}
\usepackage{multirow}
\usepackage{caption}

\usepackage[capitalize,noabbrev]{cleveref}
\usepackage[table]{xcolor}
\theoremstyle{plain}

\theoremstyle{definition}

\theoremstyle{remark}

\usepackage{enumitem}

\usepackage[utf8]{inputenc}

\usepackage{microtype}
\usepackage{inconsolata}
\usepackage{graphicx}
\title{Towards Extreme Pruning of LLMs with Plug-and-Play Mixed Sparsity}

\author{
    Chi Xu\footnotemark[1] \and  Gefei Zhang \footnotemark[1]\and  Yantong Zhu \footnotemark[1] \and Luca Benini\footnotemark[3] \and  
    Guosheng Hu\footnotemark[2] \footnotemark[4] \and  
    Yawei Li\footnotemark[3] \footnotemark[4]\and    
    Zhihong Zhang\footnotemark[1] \footnotemark[5] 
}

\footnotetext[1]{School of Informatics, Xiamen University, China.}
\footnotetext[2]{University of Bristol, England.}
\footnotetext[3]{ETH Zurich, Switzerland.}

\footnotetext[4]{Co-technical Lead}
\footnotetext[5]{Corresponding Author: zhihong@xmu.edu.cn}

\begin{document}
\maketitle
\begin{abstract}
N:M structured pruning is essential for large language models (LLMs) because it can remove less important network weights and reduce the memory and computation requirements. Existing pruning methods mainly focus on designing metrics to measure the importance of network components to guide pruning. Apart from the impact of these metrics, we observe that different layers have different sensitivities over the network performance. Thus, we propose an efficient method based on the trace of Fisher Information Matrix (FIM) to quantitatively measure and verify the different sensitivities across layers. Based on this, we propose Mixed Sparsity Pruning (MSP)  which uses a pruning-oriented evolutionary algorithm (EA) to determine the optimal sparsity levels for different layers. To guarantee fast convergence and achieve promising performance, we utilize efficient FIM-inspired layer-wise sensitivity to initialize the population of EA. In addition, our MSP can work as a plug-and-play module, ready to be integrated into existing pruning methods. Extensive experiments on LLaMA and LLaMA-2 on language modeling and zero-shot tasks demonstrate our superior performance. In particular, in extreme pruning ratio (e.g. 75\%), our method \emph{significantly}  outperforms existing methods in terms of perplexity (PPL) by orders of magnitude  (\cref{fig:contrast}).
\end{abstract}


\begin{figure}[!ht]
    \centering
    \includegraphics[width=0.6\linewidth]{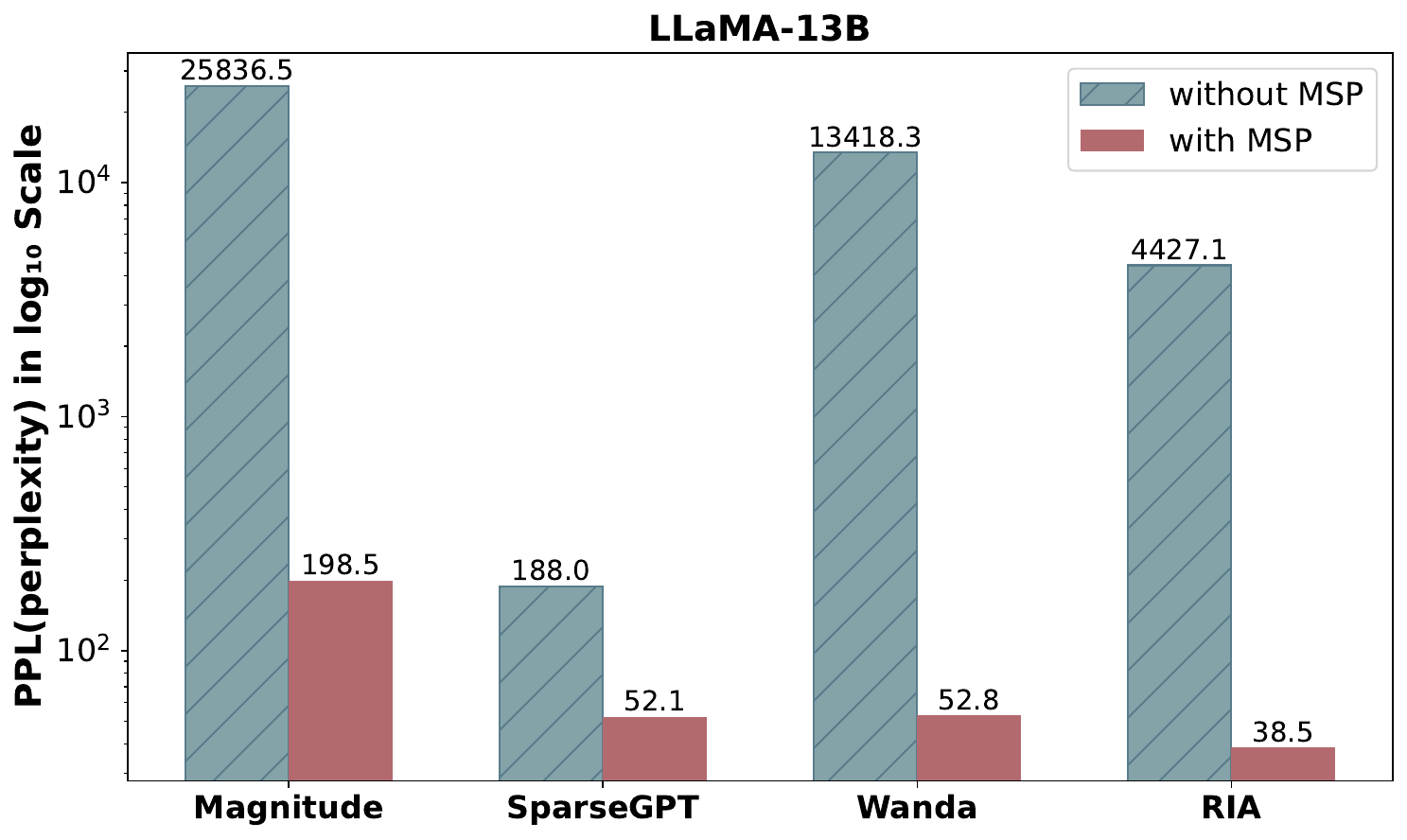}
    \vspace{-10pt}
    \caption{{Perplexity of LLaMA-13B model pruned by different methods with and without our  MSP with 75\% sparsity on WikiText dataset (lower is better). }}
    \label{fig:contrast}
    \vspace{-4pt}
\end{figure}

\section{Introduction}
\label{submission}

Despite the great success of {large language models} (LLMs) \cite{brown2020language, openai2023gpt}, the deployment of LLMs is very challenging because 
the immense parameter size requires significant memory and computational resources. One effective solution of accelerating LLMs is model pruning or model sparsity \cite{lecun1989optimal, hassibi1993optimal, lu2022learning}, which uses the inherent redundancy in LLMs to prune less important parameters without compromising performance.

Among pruning techniques, post-training pruning \cite{lu2022learning, frantar2023sparsegpt, sun2023simple}, which removes redundant parameters after model training, is particularly critical. It allows for the retention of pre-trained knowledge while significantly reducing computational complexity. This approach offers a practical pathway for optimizing LLMs, striking a balance between efficiency and effectiveness. 


Existing LLM pruning methods usually focus on designing an effective metric that determines the importance of network components in guiding pruning. For example, {Wanda \cite{sun2023simple} computes the weight importance as the element-wise product between the weight magnitude and the norm of input activations. RIA \cite{zhang2024plug} proposes relative importance and combines with activations as a new metric based on the fact that well-trained LLMs contain unique information in the input and output channels. Pruner-zero \cite{dong2024pruner} uses evolutionary algorithms to find the optimal metric for pruning.} Though these methods with various importance metrics achieve great success on sparsity ratio 50\% or smaller, their performance drops significantly on the higher ratio (e.g. $\ge$ 75\%). For example, the perplexity of LLaMA-13B pruned with Wanda \cite{sun2023simple} is 9.59 at 50\% sparsity, whereas it drastically increases to 13418.32 at 75\% sparsity.

The limitation of existing methods on very high sparsity ratio inspires us to revisit pruning methods from a fundamentally different perspective. Specifically, we observe that existing methods make an implicit assumption that the whole neural network shares an identical sparsity ratio like 50\% across different layers. Intuitively,  the ratio of unimportant weights in different layers can be different. 
Actually, existing methods have verified the soundness of our intuition. For example, the work \cite{kim2024shortened} shows that the first four and last two layers are much more important than intermediate layers of a LLM. In addition, the work HAWQ \cite{dong2019hawq} shows that different blocks in a network exhibit varying levels of sensitivity. Thus, optimizing the layer-wise sparsity ratio can potentially lead to a better pruning performance.

In this work, we first quantitatively verify the soundness of our assumption that each layer has different sensitivities. 
Existing methods \cite{dong2020hawq} measure the sensitivity of Convolutional Neural Networks (CNNs) using the trace of Hessian matrix which is unfortunately too computationally and memory-expensive for LLMs. In this work, we propose to use the trace of Fisher Information Matrix (FIM) to quantify the sensitivity to achieve efficient quantitative measurement. 
The successful verification of our assumption inspires us to propose a new 
Mixed Sparsity Pruning (MSP) method, which learns to assign different sparsity ratios to different layers, targeting the optimal pruning performance. 
Specifically,  
our MSP  uses a new pruning-oriented evolutionary algorithm (EA) to determine the optimal layer-wise sparsity levels.  To achieve fast convergence and promising search results, we initialize EA search using our proposed layer-wise sensitivity achieved by  FIM.
Once the optimal sparsity ratios across layers are searched, our MSP can work as a plug-and-play module integrated into existing pruning methods. 
%
The experimental results validate the effectiveness of the proposed MSP.



Our contributions are summarized as follows:
\begin{itemize}[nosep,leftmargin=*]

\item {We theoretically verify the soundness of the assumption that each layer has different sensitivities. To achieve this verification, we propose an efficient method based on the trace of Fisher Information Matrix to quantitatively measure the layer-wise sensitivity for LLMs.}


\item{Unlike most existing methods that focus on designing metrics of weight importance, we propose mixed sparsity pruning from a fundamentally different angle, which designs a pruning-oriented evolutionary algorithm to search the optimal layer-wise sparsity levels.}

\item {Our MSP can work as a plug-and-play module to be integrated into existing pruning methods for performance improvements.}


\item{We conduct extensive experiments on language modeling and zero-shot tasks, demonstrating the great effectiveness of our MSP. In particular,  MSP can significantly improve the performance of existing methods on extreme sparsity ratios, e.g. one or two orders of magnitude improvements on 75\% ratio, as shown in  \cref{fig:contrast}.}

\end{itemize}
\section{Related Work}

\subsection{Language Model Pruning}
Language model pruning aims to reduce their computational and memory demands while preserving performance. Recent works have introduced innovative approaches in this domain. SparseGPT \cite{frantar2023sparsegpt} formalizes the problem of pruning LLMs by solving a local layer-wise reconstruction problem, where their pruning metric and weight update procedure is inspired from Optimal Brain Surgeon (OBS) \cite{hassibi1993optimal}. Wanda \cite{kwon2022fast} streamlines the process by simplifying SparseGPT’s \cite{frantar2023sparsegpt} methodology, and explores the weight magnitude and activations as a criterion for pruning, offering a simple yet effective strategy to achieve high sparsity ratios. RIA \cite{zhang2024plug} also focuses on metrics related to weights and activations but introduces channel permutation \cite{pool2021channel} to maximize the retention of important weights under N:M sparsity. Pruner Zero \cite{dong2024pruner} employs evolutionary algorithms to discover optimal pruning metrics, providing a comprehensive framework for metric exploration. {Despite the great success of these methods, their performance drops greatly at extreme sparsity ratios, e.g. 75\%. } 

\subsection{ Sensitivity of Layers}
For CNNs, the sensitivity over the model performance varies significantly across blocks and channels \cite{liu2019metapruning, li2022revisiting, he2018amc, yu2019autoslim}. HAWQ \cite{dong2019hawq} addresses this by computing the top eigenvalues of the Hessian ($i.e.$, the second-order operator) for each block in the network to determine the appropriate quantization precision. Specifically, HAWQ \cite{dong2019hawq} ranks blocks based on their Hessian spectra and applies finer quantization to layers with larger spectra. To accommodate the large parameter size of modern models, HAWQ-V2 \cite{dong2020hawq}  refines this approach by using the average trace of the Hessian matrix, leading to more accurate quantization results.
For LLMs, not all the layers in networks exhibit the same importance, and the sensitivity varies significantly across layers. ShortGPT \cite{men2024shortgpt} takes a different approach, evaluating how much a transformer block alters the hidden states and removing layers with minimal impact.  LLM-pruner \cite{ma2023llm} identifies that the first and last layers have a profound effect on model performance, with pruning these layers causing significantly greater performance degradation than other layers. Similarly, Shortened LLaMA \cite{kim2024shortened} retains the first four and last two blocks, excluding them from pruning candidates to preserve performance. These studies collectively indicate that each layer in LLMs has different sensitivities. Thus, it might not be optimal for all the layers to share the same sparsity ratio, motivating us to explore the optimal sparsity ratios across different layers.


\section{Mixed Sparsity Pruning} 
In this work, we propose MSP, a new pruning strategy, to optimize the layer-wise sparsity ratio through a newly designed pruning-oriented evolutionary algorithm. \cref{sec:layer_sensitivity} theoretically verifies the assumption of layer-wise sensitivity with Fisher Information Matrix. 
\cref{sec:pruning-oriented evolutionary optimization} details the pruning-oriented evolutionary optimization. \cref{sec:IIEM} discusses the plug-and-play integration of MSP into existing pruning methods.

\subsection{Layer-wise Sensitivity}\label{sec:layer_sensitivity}
Our solution is based on an assumption that different layers have different sensitivities. To verify this assumption, we should quantitatively measure the sensitivity.  Existing methods HAWQ \cite{dong2019hawq} and HAWQ-V2 \cite{dong2020hawq} investigate the sensitivity of layers in CNNs for the task of model quantization. HAWQ \cite{dong2019hawq} and HAWQ-V2 \cite{dong2020hawq} use top eigenvalues and the trace of  Hessian matrix respectively. 
However, computing the Hessian matrix trace of LLMs  is computationally impractical because of the enormous parameters in  LLMs. Therefore, in this work, we propose an efficient method based on  the trace of Fisher Information Matrix, which is fast and also accurate   approximation of the Hessian matrix.

\noindent \textbf{Problem Formulation.}
Before we introduce Fisher Information Matrix, we first formulate the loss function and 2nd order Hessian matrix. 


The loss function $\mathcal{L}(\theta)$ is formulated as:
\begin{equation}
  \mathcal{L}(\theta)=\frac{1}{N} \sum_{i=1}^{N} f\left(x_{i}, y_{i}, \theta\right), 
\end{equation}
where $\theta$ $\in$ $R^{d}$ are the learnable model parameters, and $f(x,y,\theta)$ is the loss for a datum $(x,y)$ $\in$ $(X,Y)$. $N$ = $\left | X \right | $ is the cardinality of the training set. 


Existing HAWQ \cite{dong2019hawq} and HAWQ-V2 \cite{dong2020hawq} use top eigenvalues and the trace of  Hessian matrix respectively to measure the sensitivity. To calculate the Hessian matrix, 
we apply Taylor expansion to the loss function $\mathcal{L(\theta)}$ in terms of $\theta$ and approximate the loss change when pruning the weights:
\begin{equation}
    \begin{split}
        s&= \mathcal{L}(\theta-\delta\theta_{i})-\mathcal{L}(\theta)\\
        &=-\delta\theta_{i}\nabla_{\theta}\mathcal{L}+\frac{1}{2}\delta\theta_{i}^{T}(\nabla^2_{\theta}\mathcal{L})\delta\theta_{i} \\
        & =-\delta\theta_{i}^{T}g+\frac{1}{2}\delta\theta_{i}^{T}H\delta\theta_{i}=-g_{i}+\frac{1}{2}H_{ii},
    \end{split}
\end{equation}
where $\delta \theta_{i}$ means the $i^{th}$ parameter should be pruned, $g$ is the 
 gradient of the loss w.r.t model parameters $\theta$. 
After training, we can obtain $g$=$\nabla_{\theta}\mathcal{L}(\theta)$ $\approx$ $0$ \cite{lecun1989optimal} 
 and Hessian matrix
\begin{equation}
    H= \frac{\partial^2 \mathcal{L}}{\partial \theta^2}.
\end{equation}

\begin{figure}[!ht]
\centering
\begin{center}
\centerline{\includegraphics[width=0.6\columnwidth]{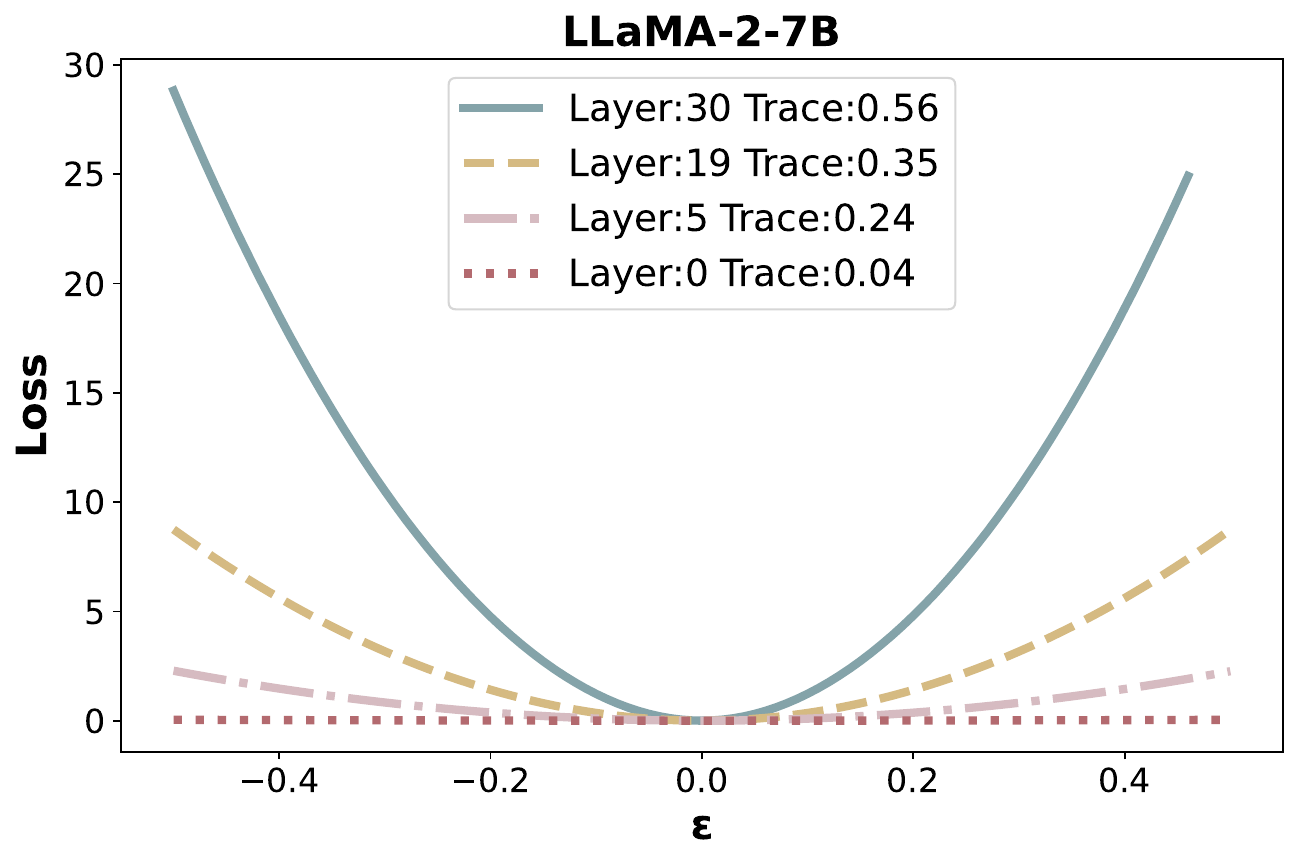}}
\caption{1-D loss landscape for different layers of LLaMA-2-7B on C4 datasets \cite{raffel2020exploring}. The landscape is plotted by perturbing model weights along the trace of Fisher Information Matrix of each layer, with a magnitude of $\epsilon$($i.e.$, $\epsilon=0$ corresponds to no perturbation).} 
\label{fig:loss}
\end{center}
\vspace{-24pt}
\end{figure}
\noindent \textbf{Fisher Information Matrix.}
{However, computing the aforementioned Hessian matrix trace is computationally impractical due to the numerous parameters of LLMs.} Considering the trace of Hessian matrix, we only need compute the sum of the diagonal entry $H_{ii}$ of Hessian matrix $H$. So we apply the {Fisher Information Matrix} \cite{kwon2022fast, liu2021group, theis2018faster} to approximate the Hessian matrix:
\begin{equation}
\label{equ:transfer}
    \begin{split}
        H_{ii}&=\frac{\partial^2\mathcal{L}}{\partial \theta_{i}^2}=\frac{1}{N}\sum_{n=1}^{N}\frac{\partial^2\mathcal{L}_{n}}{\partial \theta_{i}^2} \approx-\frac{\partial^2 \mathbb{E}[logp(y|x)]}{\partial\theta_{i}^{2}}\\
        &=\mathbb{E}\left [-\frac{\partial logp(y|x)}{\partial \theta_{i}}\right]^2\approx\frac{1}{N}\sum_{n=1}^{N}\left(\frac{\partial \mathcal{L}_{n}}{\partial \theta_{i}}\right)^2.
    \end{split}
\end{equation}
Thus, we employ Fisher information matrix to transform second-order derivative to the square of first-order derivative, significantly reducing the computation complexity.

\noindent \textbf{Sensitivity of Layers.}
We computed the Hessian traces for each layer and visualized the 1-D loss landscapes of several representative layers, as shown in \cref{fig:loss}. Under identical perturbations, layers with larger trace values exhibited greater loss increases, indicating higher sensitivity. Clearly, different layers have very different sensitivities, which inspires us to assign different sparsity ratios to different layers in \cref{sec:pruning-oriented evolutionary optimization}.


\subsection{Pruning-oriented Evolution}\label{sec:pruning-oriented evolutionary optimization}

In this section, we detail our MSP strategy using evolutionary optimization framework as shown in \cref{alg:overall_evolutionary_algorithm}. We design a new evolutionary algorithm specifically for solving our pruning problem. In the evolutionary algorithm, each individual represents a unique set of sparsity levels across layers, where each sparsity level corresponds to the proportion of weights to be pruned from a specific layer of a large language model. During the evolution process, the individuals are evaluated by a fitness function, and the top-performing individuals are selected as parents to generate offsprings. The offsprings undergo crossover (\cref{alg:crossover} in \cref{asubsec:crossover}) and mutation (\cref{alg:mutation} in \cref{asubsec:mutation})  following predefined crossover and mutation rates, introducing variability in the population. A fixed target sparsity level is predefined and the average pruning sparsity ratio across each individual is constrained to maintain this target sparsity ratio, ensuring adherence to the global pruning requirements. 


\begin{algorithm}[!ht]
\caption{Evolutionary Pruning Algorithm for Language Models}
    \label{alg:overall_evolutionary_algorithm}
    \begin{algorithmic}
        \STATE {\bfseries Input:} Target sparsity ratio $N:M$, Mutation rate $\mu$, Population size $P$, Generations $G$
        \STATE {\bfseries Output:} Optimal sparsity configuration $ind^*$ and corresponding perplexity $ppl^*$
    \end{algorithmic}  
    \begin{algorithmic}[1]
        \STATE Initialize population $popu$ of size $P$ with  individuals satisfying $N:M$ sparsity constraints\FOR{$g = 1$ {\bfseries to} $G$}
         \STATE Evaluate fitness for each individual in $popu$
         \STATE $parents \gets \text{select}(popu)$
         \STATE $offspring \gets \text{crossover}(parents)$
         \STATE $offspring \gets \text{mutate}(offspring, \mu)$
         \STATE $popu \gets \text{update\_population}(offspring)$
      \ENDFOR
        \STATE Evaluate final population to find the individual $ind^*$ with the lowest perplexity $ppl^*$
        \STATE \textbf{return} {$ind^*$, $ppl^*$}
   \end{algorithmic}
\end{algorithm}

\begin{figure}[!t]
    \centering
    \includegraphics[width=0.6\linewidth]{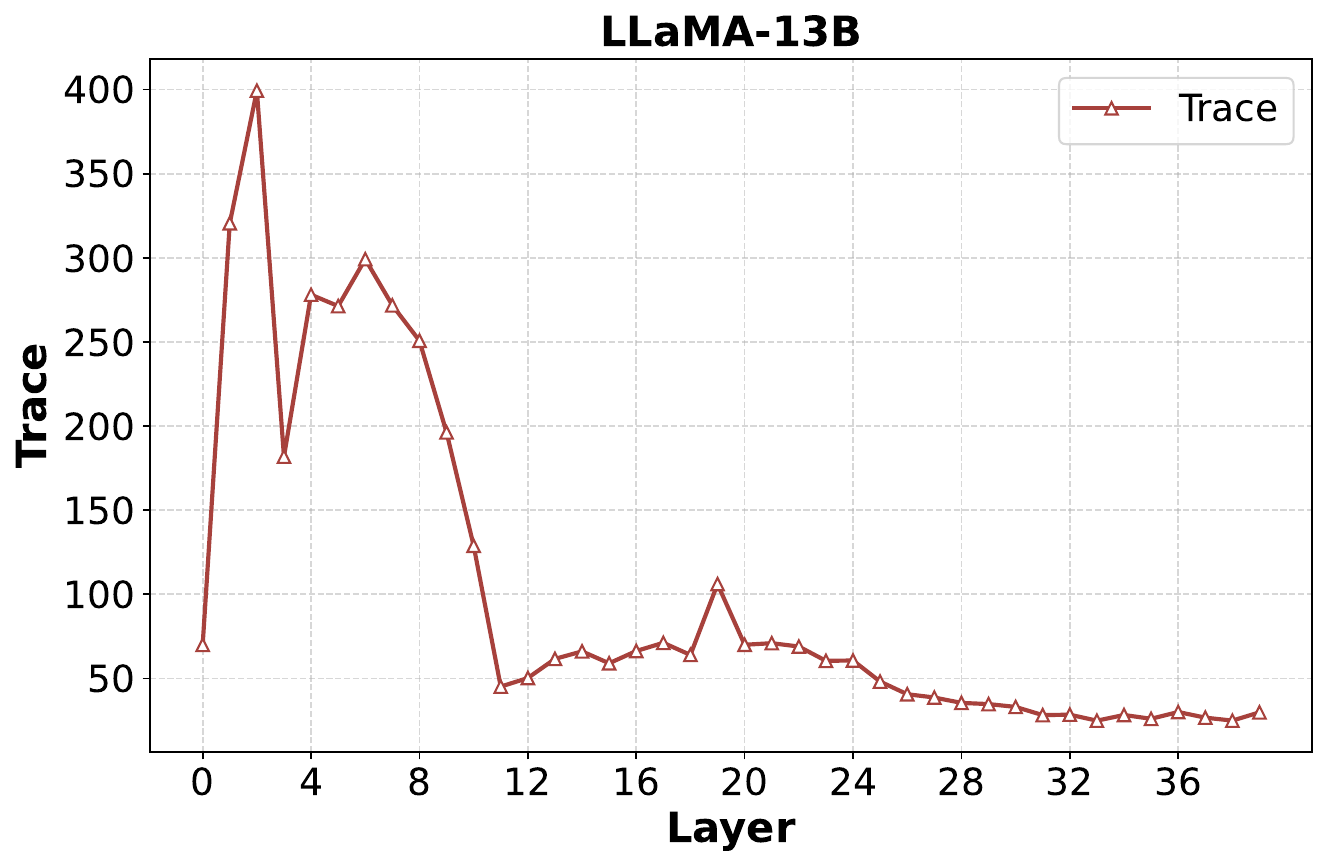}
    \vspace{-10pt}
    \caption{{The trace of Hessian matrix of LLaMA-13B.}}
    \label{fig:trace}
    \vspace{-4pt}
\end{figure}

\noindent \textbf{Sensitivity-informed Initialization of Population.} 
We analyze and calculate the layer-wise sensitivity of LLMs using our proposed evaluation metric in \cref{sec:layer_sensitivity}. We empirically find that the first few layers of LLMs are more sensitive than the deeper layers.  For example, as shown in \cref{fig:trace}, the trace values of the first 8 layers of a LLaMA-13B model (roughly front ($\frac{1}{5}$) layers) are significantly higher than the deeper layers. Similar observations can also be found at  \cite{kim2024shortened, ma2023llm}. Thus, the sparsity ratios of the first few layers, roughly first ($\frac{1}{5}$) layers, are expected to be lower than the other layers to achieve promising performance. Motivated by this, we initialize the individuals with the constraint that the sparsity ratios of the first  ($\frac{1}{5}$) layers are smaller than the others. This constraint applies only during initialization and does not affect the search process. The exact initialization process can be found at \cref{alg:sensitivity_init} of \cref{asubsec:sensitivity-inspired_initialization}. In \cref{sec:ablation_study}, we empirically show that the sensitivity-informed initialization of population outperforms random initialization.  Finally, we present one example of this initialization. Given a network with only five layers and a target pruning ratio of 75\% (N:M=3:4), an individual is possibly initialized as [50\% (2:4), 75\% (3:4), 100\% (4:4), 75\% (3:4), 75\% (3:4)].


\noindent \textbf{Pruning-Oriented Crossover.} 
Crossover is performed between two parent individuals to produce two child individuals. A random crossover point splits the parents into segments, which are then combined to form the children. Adjustments are made by progressively increasing or decreasing the sparsity level of a random selected layer within the allowed range from 0:M to M:M until the average sparsity equals the target sparsity again. More details can be found at \cref{alg:crossover} (\cref{asubsec:crossover}).

\noindent \textbf{Pruning-Oriented Mutation.}
{After crossover, mutation is applied to introduce further variability. In \cref{alg:mutation} of \cref{asubsec:mutation}, two layers within an individual are randomly selected as mutation points since mutating only a single layer would disrupt the target average sparsity ratio. Pruning ratios of the two layers are mutated according to a pre-defined mutation rate. If the constraint that the average pruning ratio across all layers is equal to the target sparsity is violated, the original sparsity ratios are restored, and the mutation step is repeated until a valid individual is generated.}

\noindent \textbf{Perplexity as Fitness.} In our evolutionary algorithm, perplexity (ppl) is used as the fitness score to evaluate the quality of an individual. The goal of our pruning-oriented evolutionary algorithm is to minimize perplexity while ensuring that the model’s sparsity constraints are maintained  throughout the process. 

\subsection{Plug-and-play Integration}\label{sec:IIEM}
Once the evolutionary process is complete, the optimal sparsity levels for each layer are determined. Based on the layer-wise sparsity, we can use any metric, which measures the importance of network components, to conduct pruning. In this way, our MSP can work as a plug-and-play module which can easily be integrated into those existing methods that design importance metrics. As will be demonstrated, the performance of existing pruning methods can be elegantly improved by integrating the proposed MSP. 

\begin{table*}[!htb]
\centering
\caption{WikiText perplexity(PPL) of pruned LLMs with 75\% sparsity. When integrated with our MSP, SOTA pruning methods achieve significantly lower perplexity.}
\label{table:msp_ppl_main_result_r75}
\vspace{-2pt}
\resizebox{1\columnwidth}{!}{
\begin{tabular}{ccccccccc}
\toprule[0.1em]
                         &                            & \multicolumn{4}{c}{LLaMA}                                             & \multicolumn{3}{c}{LLaMA-2}                         \\
\multirow{-2}{*}{Method} & \multirow{-2}{*}{Sparsity} & 7B               & 13B             & 30B             & 65B            & 7B               & 13B             & 70B            \\ \hline
Dense                    & -                          & 5.68             & 5.09            & 4.10            & 3.53           & 5.12             & 4.57            & 3.12           \\ \hline
Magnitude                & 75\%(3:4)                  & 102146.91        & 25836.51        & 13011.61        & 13130.95       & 30279.95         & 15711.99        & 20738.73       \\
\rowcolor[HTML]{EFEFEF} 
Magnitude + MSP          & 75\%(M=4)                  & \textbf{5531.29} & \textbf{198.51} & \textbf{101.50} & \textbf{37.57} & \textbf{3068.93} & \textbf{461.97} & \textbf{43.96} \\
SparseGPT                & 75\%(3:4)                  & 319.9            & 187.99          & 91.01           & 74.74          & 134.36           & 116.58          & 59.49          \\
\rowcolor[HTML]{EFEFEF} 
SparseGPT + MSP          & 75\%(M=4)                  & \textbf{204.61}  & \textbf{52.13}  & \textbf{24.36}  & \textbf{21.22} & \textbf{134.26}  & \textbf{70.88}  & \textbf{19.03} \\
Wanda                    & 75\%(3:4)                  & 7293.53          & 13418.32        & 4489.07         & 2498.61        & 3930.21          & 3348.26         & 374.03         \\
\rowcolor[HTML]{EFEFEF} 
Wanda + MSP              & 75\%(M=4)                  & \textbf{181.26}  & \textbf{52.76}  & \textbf{36.57}  & \textbf{23.43} & \textbf{139.94}  & \textbf{94.18}  & \textbf{22.72} \\ \hline
Magnitude                & 75\%(6:8)                  & 96129.87         & 44160.89        & 6101.86         & 7210.08        & 40190.01         & 10002.21        & 17257.87       \\
\rowcolor[HTML]{EFEFEF} 
Magnitude + MSP          & 75\%(M=8)                  & \textbf{4330.54} & \textbf{128.11} & \textbf{53.59}  & \textbf{28.99} & \textbf{1557.09}     & \textbf{90.72}  & \textbf{34.24} \\
SparseGPT                & 75\%(6:8)                  & 177.61           & 92.35           & 59.11           & 39.76          & 76.35            & 60.39           & 33.20          \\
\rowcolor[HTML]{EFEFEF} 
SparseGPT + MSP          & 75\%(M=8)                  & \textbf{87.11}   & \textbf{31.95}  & \textbf{20.12}  & \textbf{17.21} & \textbf{51.02}     & \textbf{38.61}  & \textbf{14.66} \\
Wanda                    & 75\%(6:8)                  & 5288.80          & 3736.05         & 1084.95         & 822.97         & 1839.8           & 1258.42         & 151.4          \\
\rowcolor[HTML]{EFEFEF} 
Wanda + MSP              & 75\%(M=8)                  & \textbf{158.03}  & \textbf{35.66}  & \textbf{24.93}  & \textbf{17.98} & \textbf{115.63}  & \textbf{46.08}  & \textbf{15.70} \\ \hline
Magnitude                & 75\%(12:16)                & 65567.26         & 13800.68        & 4658.58         & 5828.06        & 21597.39         & 7778.01         & 5795.37        \\
\rowcolor[HTML]{EFEFEF} 
Magnitude + MSP          & 75\%(M=16)                 & \textbf{909.01}  & \textbf{115.94} & \textbf{50.98}  & \textbf{27.03} & \textbf{1884.94}     & \textbf{78.55}  & \textbf{31.94} \\
SparseGPT                & 75\%(12:16)                & 100.04           & 67.91           & 44.69           & 30.57          & 62.41            & 44.36           & 22.95          \\
\rowcolor[HTML]{EFEFEF} 
SparseGPT + MSP          & 75\%(M=16)                 & \textbf{62.28}   & \textbf{28.66}  & \textbf{17.98}  & \textbf{15.75} & \textbf{42.51}   & \textbf{31.18}  & \textbf{13.20} \\
Wanda                    & 75\%(12:16)                & 2175.68          & 1218.78         & 471.79          & 321.47         & 1406.78          & 351.19          & 73.78          \\
\rowcolor[HTML]{EFEFEF} 
Wanda + MSP              & 75\%(M=16)                 & \textbf{88.07}   & \textbf{34.51}  & \textbf{21.18}  & \textbf{15.69} & \textbf{68.52}   & \textbf{37.65}  & \textbf{13.64} \\ \bottomrule[0.1em]
\end{tabular}
}
\end{table*}

\begin{table*}[!htb]
\centering
\caption{Mean zero-shot accuracies (\%) of pruned LLaMA and LLaMA-2 models at the sparsity of 75\%. When integrated with our MSP, SOTA pruning methods achieve performance much better than their original implementations.}
\vspace{-2pt}
\label{msp_tasks_acc_main_result_r75}
\resizebox{0.8\columnwidth}{!}{
\begin{tabular}{ccccccccc}
\toprule[0.1em]
                         &                            & \multicolumn{4}{c}{LLaMA}                                         & \multicolumn{3}{c}{LLaMA-2}                      \\
\multirow{-2}{*}{Method} & \multirow{-2}{*}{Sparsity} & 7B             & 13B            & 30B            & 65B            & 7B             & 13B            & 70B            \\ \hline
Dense                    & -                          & 60.01          & 62.60          & 65.37          & 66.99          & 59.68          & 63.02          & 67.13          \\ \hline
Magnitude                & 75\%(3:4)                  & 34.99          & 31.98          & 32.84          & 32.91          & 32.15          & 33.32          & 33.15          \\
\rowcolor[HTML]{EFEFEF} 
Magnitude + MSP          & 75\%(M=4)                  & \textbf{35.49} & \textbf{39.72} & \textbf{40.17} & \textbf{41.82} & \textbf{35.23} & \textbf{35.76} & \textbf{41.47} \\
SparseGPT                & 75\%(3:4)                  & 32.53          & 34.39          & 36.35          & 38.01          & 32.13          & 33.25          & 37.47          \\
\rowcolor[HTML]{EFEFEF} 
SparseGPT + MSP          & 75\%(M=4)                  & \textbf{34.95} & \textbf{39.10} & \textbf{41.66} & \textbf{46.04} & \textbf{34.01} & \textbf{39.29} & \textbf{44.42} \\
Wanda                    & 75\%(3:4)                  & 32.63          & 31.84          & 34.08          & 32.06          & 32.08          & 32.32          & 32.79          \\
\rowcolor[HTML]{EFEFEF} 
Wanda + MSP              & 75\%(M=4)                  & \textbf{36.75} & \textbf{38.93} & \textbf{40.89} & \textbf{45.67} & \textbf{33.80} & \textbf{38.53} & \textbf{43.33} \\ \hline
Magnitude                & 75\%(6:8)                  & 33.87          & 33.36          & 32.04          & 33.37          & 32.36          & 32.65          & 32.77          \\
\rowcolor[HTML]{EFEFEF} 
Magnitude + MSP          & 75\%(M=8)                  & \textbf{36.55} & \textbf{39.76} & \textbf{41.17} & \textbf{45.76} & \textbf{35.08}   & \textbf{37.66} & \textbf{42.87} \\
SparseGPT                & 75\%(6:8)                  & 32.95          & 36.64          & 39.49          & 40.93          & 33.13          & 36.55          & 41.08          \\
\rowcolor[HTML]{EFEFEF} 
SparseGPT + MSP          & 75\%(M=8)                  & \textbf{37.46} & \textbf{40.50} & \textbf{43.28} & \textbf{48.21} & \textbf{36.76}   & \textbf{40.28} & \textbf{46.72} \\
Wanda                    & 75\%(6:8)                  & 33.53          & 32.03          & 33.66          & 32.11          & 32.46          & 32.41          & 35.15          \\
\rowcolor[HTML]{EFEFEF} 
Wanda + MSP              & 75\%(M=8)                  & \textbf{37.75} & \textbf{40.32} & \textbf{43.33} & \textbf{48.18} & \textbf{38.21} & \textbf{38.49} & \textbf{46.28} \\ \hline
Magnitude                & 75\%(12:16)                & 32.63          & 33.46          & 33.61          & 34.35          & 32.80          & 32.88          & 34.33          \\
\rowcolor[HTML]{EFEFEF} 
Magnitude + MSP          & 75\%(M=16)                 & \textbf{37.48} & \textbf{38.74} & \textbf{42.16} & \textbf{45.95} & \textbf{33.78} & \textbf{40.45} & \textbf{43.81} \\
SparseGPT                & 75\%(12:16)                & 35.66          & 37.45          & 40.94          & 43.93          & 36.39          & 38.28          & 44.76          \\
\rowcolor[HTML]{EFEFEF} 
SparseGPT + MSP          & 75\%(M=16)                 & \textbf{38.20} & \textbf{41.40} & \textbf{43.43} & \textbf{50.09} & \textbf{38.18}   & \textbf{40.83} & \textbf{47.81} \\
Wanda                    & 75\%(12:16)                & 32.44          & 35.65          & 36.37          & 34.98          & 32.47          & 32.04          & 36.35          \\
\rowcolor[HTML]{EFEFEF} 
Wanda + MSP              & 75\%(M=16)                 & \textbf{38.73} & \textbf{40.36} & \textbf{44.18} & \textbf{49.32} & \textbf{36.09} & \textbf{38.95} & \textbf{46.83} \\ \bottomrule[0.1em]
\end{tabular}
}
\vspace{-2pt}
\end{table*}

\section{Experiments}
\subsection{Models and Experiment Details}
\noindent \textbf{Models and Evaluation.} We evaluate the pruned models on language modeling and zero-shot tasks. For language modeling, we evaluate MSP on the most widely adopted LLM models: LLaMA 7B/13B/30B/65B \cite{touvron2023LLaMA}, LLaMA-2 7B/13B/70B \cite{touvron2023LLaMA2}, and OPT 125m/350m/1.3B/2.7B/6.7B/13B\cite{zhang2022opt} (\cref{asubsec:opt_results}).
For zero-shot evaluation, we conduct seven tasks from EleutherAI LM Harness \cite{gao2021framework}. Following previous works on LLM compression \cite{xiao2023smoothquant}, we evaluate the perplexity on the held-out WikiText \cite{merity2016pointer} validation dataset.

\noindent \textbf{Baselines.} We conduct experiments on three prior pruning methods including magnitude pruning\cite{han2015deep}, SparseGPT\cite{frantar2023sparsegpt}, and Wanda\cite{sun2023simple}, and compare  those methods before and after the integration of our MSP. 

\noindent \textbf{Sparsity.} 
We mainly evaluate four types of target sparsity: structured 2:4, structured 3:4, structured 6:8, and structured 12:16. We adopt N:M sparsity across all linear layers in our pruning settings. However, unlike existing pruning methods, MSP assigns different sparsity levels to different linear layers rather than applying a uniform ratio. To ensure a fair comparison with other pruning methods, in our MSP, we set the same M as the target pruning of the other methods while searching for the optimal N for different layers.


\noindent \textbf{Evolution Settings.} The evolutionary algorithm is initialized with 20 individuals and proceeds with 20 iterations. 
Half of the best-performing individuals in each iteration are selected as the parents for crossover. 
The mutation rate is set to 0.5. 
Experiments are done with 2 NVIDIA A100 GPUs.

\subsection{Language Modeling}\label{sec:language_modeling}

In \cref{table:msp_ppl_main_result_r75}, we report the perplexity of pruned LLaMA and LLaMA-2 models under 75\% sparsity. We evaluate models pruned with 3:4, 6:8, and 12:16 structured sparsity using Magnitude \cite{han2015deep}, SparseGPT \cite{frantar2023sparsegpt} and Wanda \cite{sun2023simple}, as well as all the aforementioned methods using our MSP approach under the same 75\% sparsity (the value of "M" in \cref{table:msp_ppl_main_result_r75} remains consistent with the target sparsity N:M). By applying the mixed sparsity ratios across layers determined by our MSP, existing pruning methods achieve significant improvements, with perplexity reduced by 8–10 times and up to 254 times in certain cases. 
For instance, with our MSP, Wanda \cite{sun2023simple} achieves a significantly lower perplexity (52.76) compared to its original performance (13418.32) under the same 75\% sparsity level on LLaMA-13B. These results indicate the existence of precise and effective sparse sub-networks for LLaMA models.

\begin{table*}[!htb]
\centering
\caption{Accuracies (\%) of LLaMA-13B for 7 zero-shot tasks with 75\%(M=4) sparsity. When integrated with MSP, SOTA pruning methods achieve improved performance.}
\label{msp_tasks_acc_result_r75}
\resizebox{1\columnwidth}{!}{
\begin{tabular}{ccccccccc}
\toprule[0.1em]
Task          & BoolQ          & RTE            & HellaSwag      & WinoGrande     & ARC-e          & ARC-c          & OBQA           & AVG            \\ \hline
Dense           & 77.91          & 70.75          & 59.93          & 72.61          & 77.36          & 46.42          & 33.20          & 62.60          \\ \hline
Magnitude       & 38.59          & 51.26          & 25.79          & 48.78          & 25.46          & 21.16          & 12.80          & 31.98          \\
\rowcolor[HTML]{EFEFEF} 
Magnitude + MSP & \textbf{62.08} & \textbf{54.51} & \textbf{31.12} & \textbf{51.86} & \textbf{36.53} & \textbf{22.95} & \textbf{19.00} & \textbf{39.72} \\
SparseGPT       & 51.80          & 52.71          & 27.93          & 48.70          & 28.41          & 17.41          & 13.80          & 34.39          \\
\rowcolor[HTML]{EFEFEF} 
SparseGPT + MSP & \textbf{58.38} & 52.71          & \textbf{32.76} & \textbf{51.78} & \textbf{39.65} & \textbf{22.44} & \textbf{16.00} & \textbf{39.10} \\
Wanda           & 37.83          & 46.93          & 25.78          & 50.36          & 26.26          & 20.73          & 15.00          & 31.84          \\
\rowcolor[HTML]{EFEFEF} 
Wanda + MSP     & \textbf{59.79} & \textbf{52.71} & \textbf{32.23} & \textbf{51.46} & \textbf{39.39} & \textbf{21.33} & \textbf{15.60} & \textbf{38.93} \\ \hline
\end{tabular}
}
\vspace{-2pt}
\end{table*}

\subsection{Zero-shot Tasks}\label{sec:zero-shot tasks}
To assess the generalizability of MSP, we evaluate its performance on seven common-sense tasks including BoolQ \cite{clark2019boolq}, RTE \cite{wang2018glue}, HellaSwag \cite{zellers2019hellaswag}, WinoGrande \cite{sakaguchi2021winogrande}, ARC \cite{clark2019boolq}, and OBQA \cite{mihaylov2018can} from the EleutherAI lm-evaluation-harness benchmark \cite{gao2021framework} in a zero-shot setting. The results for the seven tasks with 75\% (M=4) sparsity are shown in \cref{msp_tasks_acc_result_r75}. To avoid the variability that might exhibited on individual tasks, we compute the average accuracies across all tasks and settings in \cref{msp_tasks_acc_main_result_r75}.
The results show that when MSP is integrated, the performance of existing pruning methods is significantly improved in most settings at a high sparsity ratio of 75\%. This demonstrates the effectivenss and generalizablity of the our MSP.


\subsection{Ablation Study} 
\label{sec:ablation_study}

\begin{figure}[!ht]
    \centering
    \includegraphics[width=0.6\linewidth]{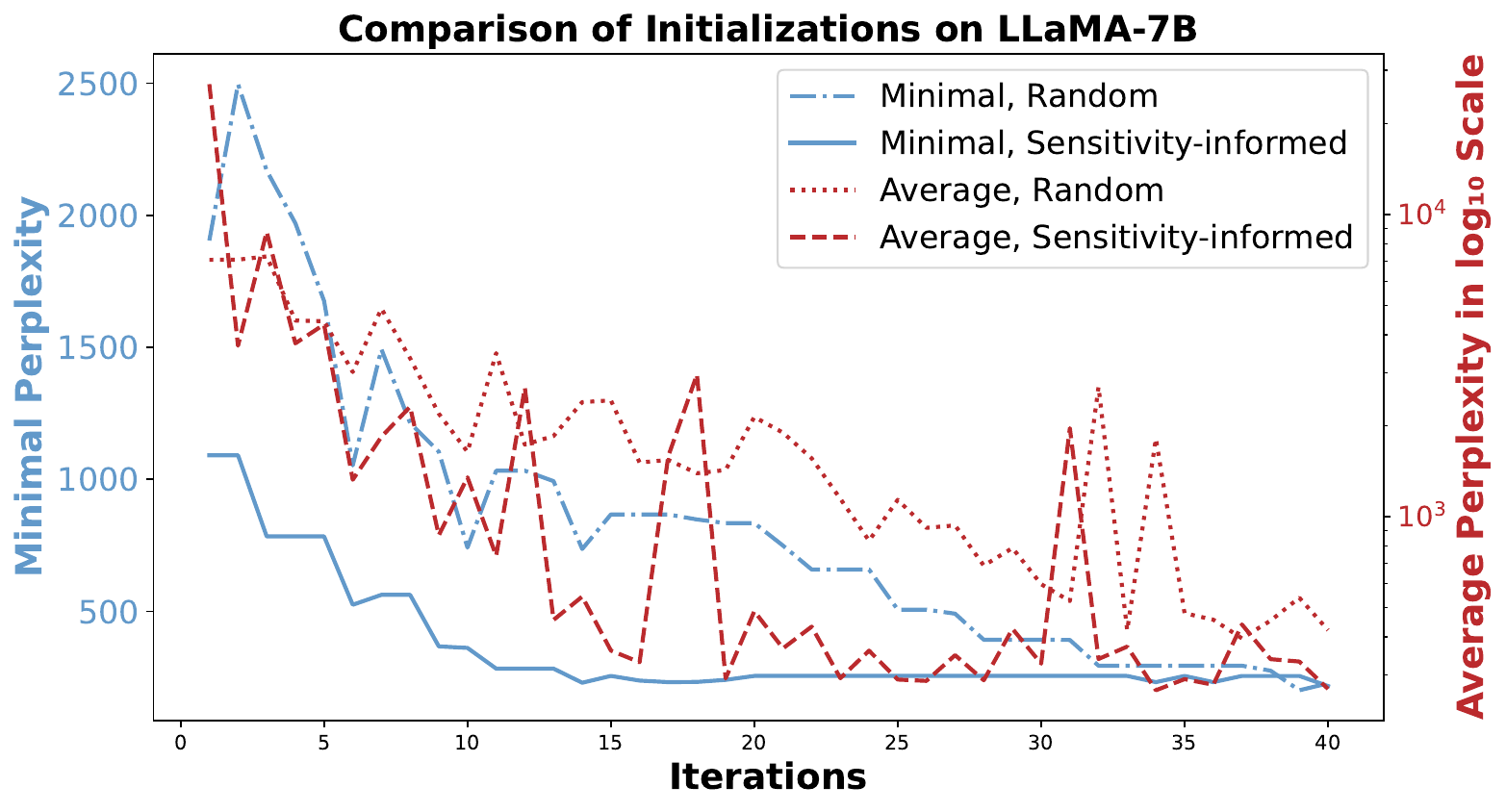}
    \vspace{-10pt}
    \caption{Comparison between random initialization and sensitivity-informed initialization on LLaMA-7B.}
    \label{fig:ablation_iterations}
    \vspace{-4pt}
\end{figure}
\noindent \textbf{Random Initialization vs. Sensitivity-informed Initialization.} To evaluate the effectiveness of sensitivity-informed initialization, we conduct experiments on LLaMA-7B in \cref{fig:ablation_iterations} to compare it with the widely used random initialization.
The results clearly show that sensitivity-informed initialization leads to faster convergence than random initialization (\cref{asubsec:random_initialization}). Therefore, all our experiments use sensitivity-informed initialization due to its fast convergence. 

\begin{figure}[!ht]
    \centering
    \includegraphics[width=0.6\linewidth]{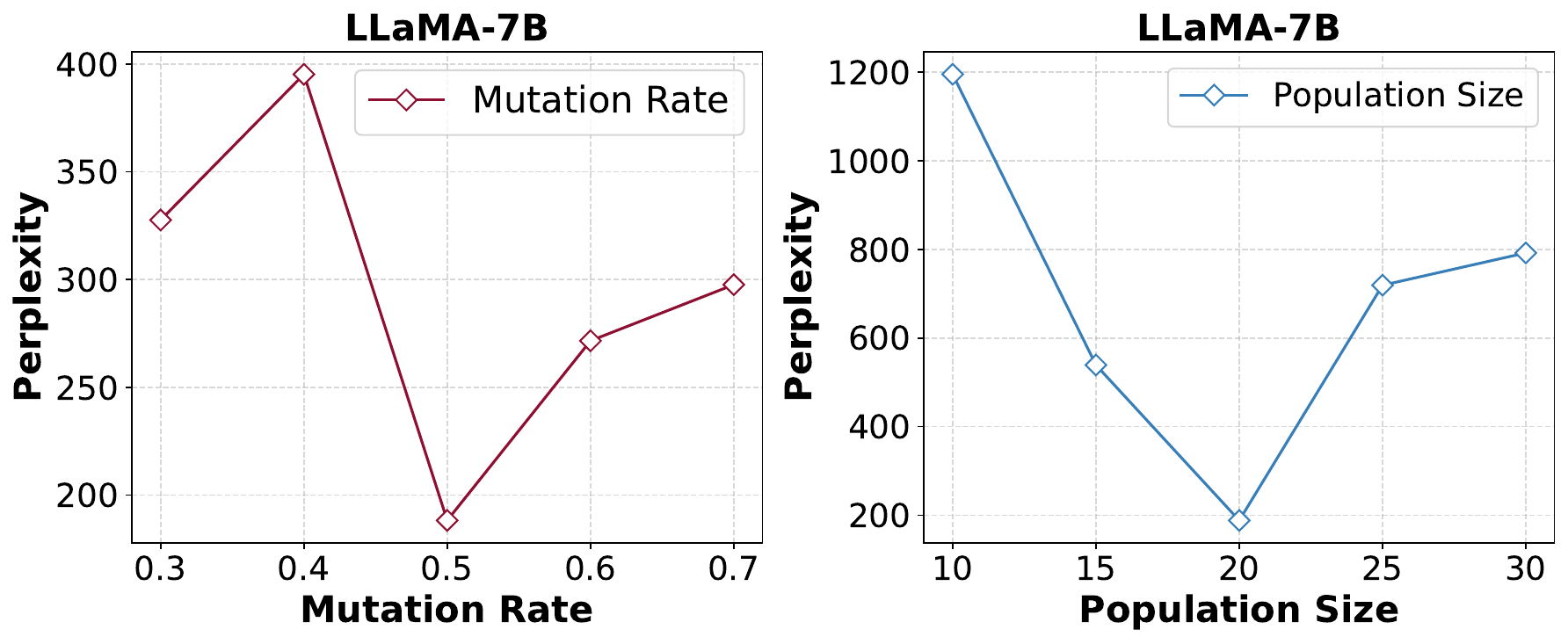}
    \caption{Ablation study on mutation rates (left) and population sizes (right) with LLaMA-7B model.}
    \label{fig:ablation_mutation_pop}
    \vspace{-4pt}
\end{figure}

\noindent \textbf{Evolution Parameters.} \cref{fig:ablation_iterations} illustrates the impact of evolutionary iterations on the minimal perplexity and the average perplexity of individuals in every generation. We evaluate iterations ranging from 1 to 40. The results show a clear trend: as the number of iterations increases, the perplexity of the optimal individual decreases. However, with sensitivity-informed initialization, beyond 20 iterations, the minimal and average perplexity of populations stabilizes and shows minimal change.  \cref{fig:ablation_mutation_pop} represents how the size of the population and mutation rate impact the results of experiments. Both excessively small and large population sizes or mutation rates negatively affect perplexity. Therefore, we set the mutation rate to 0.5 and the population size to 20.

\begin{figure}[!ht]
    \centering
    \includegraphics[width=0.6\linewidth]{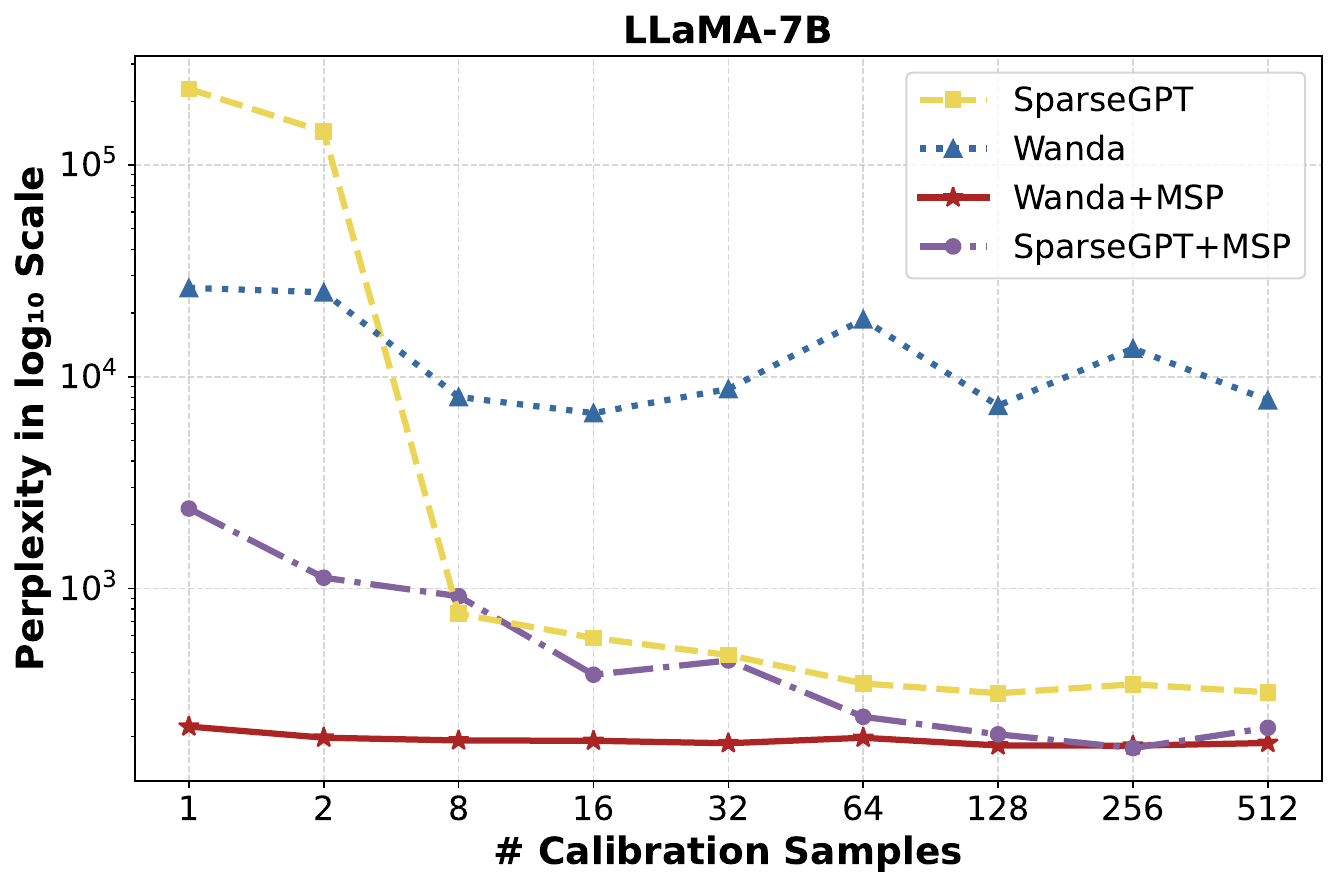}
    \vspace{-10pt}
    \caption{Perplexity with different calibration samples on pruned LLaMA-7B.}
    \label{fig:ablation_nsamples}
\end{figure}

\begin{table*}[ht]
\vspace{-0.5cm}
        \centering
        \caption{Extreme sparsity analysis on pruned LLaMA-13B. MSP consistently improves the performance of baseline methods under extreme sparsity levels.}
        \vspace{-2pt}
        \label{table:extremely_pruning_results}
        \resizebox{1\columnwidth}{!}{
        \begin{tabular}{ccccccccccc}
            \toprule[0.1em]
            \multicolumn{11}{c}{LLaMA-13B dense baseline: 5.09}  \\ \midrule[0.1em]
            & \multicolumn{3}{c}{50\%}                                                                                                 & \multicolumn{2}{c}{62.5\%}                                                      & \multicolumn{3}{c}{75\%}                                                                                                    & \multicolumn{2}{c}{87.50\%}                                                           \\ \cline{2-11} 
            
            \multirow{-2}{*}{Method}                & 2:4                                    & 4:8                                    & 8:16                                   & 5:8                                    & 10:16                                  & 3:4                                     & 6:8                                     & 12:16                                   & 7:8                                       & 14:16                                     \\ \hline
            Magnitude& 18.32                                  & 13.72                                  & 12.65                                  & 607.68                                 & 270.09                                 & 25836.51                                & 44160.89                                & 13800.68                                & 84946.19                                  & 59483.72                                  \\
            
            \cellcolor[HTML]{EFEFEF}Magnitude + MSP & \cellcolor[HTML]{EFEFEF}\textbf{15.66} & \cellcolor[HTML]{EFEFEF}\textbf{11.97} & \cellcolor[HTML]{EFEFEF}\textbf{11.93} & \cellcolor[HTML]{EFEFEF}\textbf{32.26} & \cellcolor[HTML]{EFEFEF}\textbf{30.15} & \cellcolor[HTML]{EFEFEF}\textbf{198.51} & \cellcolor[HTML]{EFEFEF}\textbf{128.11} & \cellcolor[HTML]{EFEFEF}\textbf{115.94} & \cellcolor[HTML]{EFEFEF}\textbf{29366.31} & \cellcolor[HTML]{EFEFEF}\textbf{25902.26} \\
            
            SparseGPT                               & 9.01                                   & 7.38                                   & 6.75                                   & 16.36                                  & 13.26                                  & 187.99                                  & 92.35                                   & 67.91                                   & 3305.37                                   & 2156.88                                   \\
            
            \cellcolor[HTML]{EFEFEF}SparseGPT + MSP & \cellcolor[HTML]{EFEFEF}\textbf{8.05}  & \cellcolor[HTML]{EFEFEF}\textbf{7.12}  & \cellcolor[HTML]{EFEFEF}\textbf{6.66}  & \cellcolor[HTML]{EFEFEF}\textbf{11.70} & \cellcolor[HTML]{EFEFEF}\textbf{10.29} & \cellcolor[HTML]{EFEFEF}\textbf{52.13}  & \cellcolor[HTML]{EFEFEF}\textbf{31.95}  & \cellcolor[HTML]{EFEFEF}\textbf{28.66}  & \cellcolor[HTML]{EFEFEF}\textbf{1005.33}  & \cellcolor[HTML]{EFEFEF}\textbf{866.65}   \\
            
            Wanda                                   & 9.59                                   & 7.40                                   & 6.73                                   & 33.67                                  & 17.41                                  & 13418.32                                & 3736.05                                 & 1218.78                                 & 72371.68                                  & 16072.48                                  \\
            
            \cellcolor[HTML]{EFEFEF}Wanda + MSP     & \cellcolor[HTML]{EFEFEF}\textbf{8.02}  & \cellcolor[HTML]{EFEFEF}\textbf{6.96}  & \cellcolor[HTML]{EFEFEF}\textbf{6.49}  & \cellcolor[HTML]{EFEFEF}\textbf{12.21} & \cellcolor[HTML]{EFEFEF}\textbf{10.48} & \cellcolor[HTML]{EFEFEF}\textbf{52.76}  & \cellcolor[HTML]{EFEFEF}\textbf{35.66}  & \cellcolor[HTML]{EFEFEF}\textbf{34.51}  & \cellcolor[HTML]{EFEFEF}\textbf{1421.25}  & \cellcolor[HTML]{EFEFEF}\textbf{1160.59}  \\ \bottomrule[0.1em]
        \end{tabular}
        }
        \vspace{-4pt}

\end{table*}

\begin{figure}[!ht]
   \centering
    \includegraphics[width=0.5\linewidth]{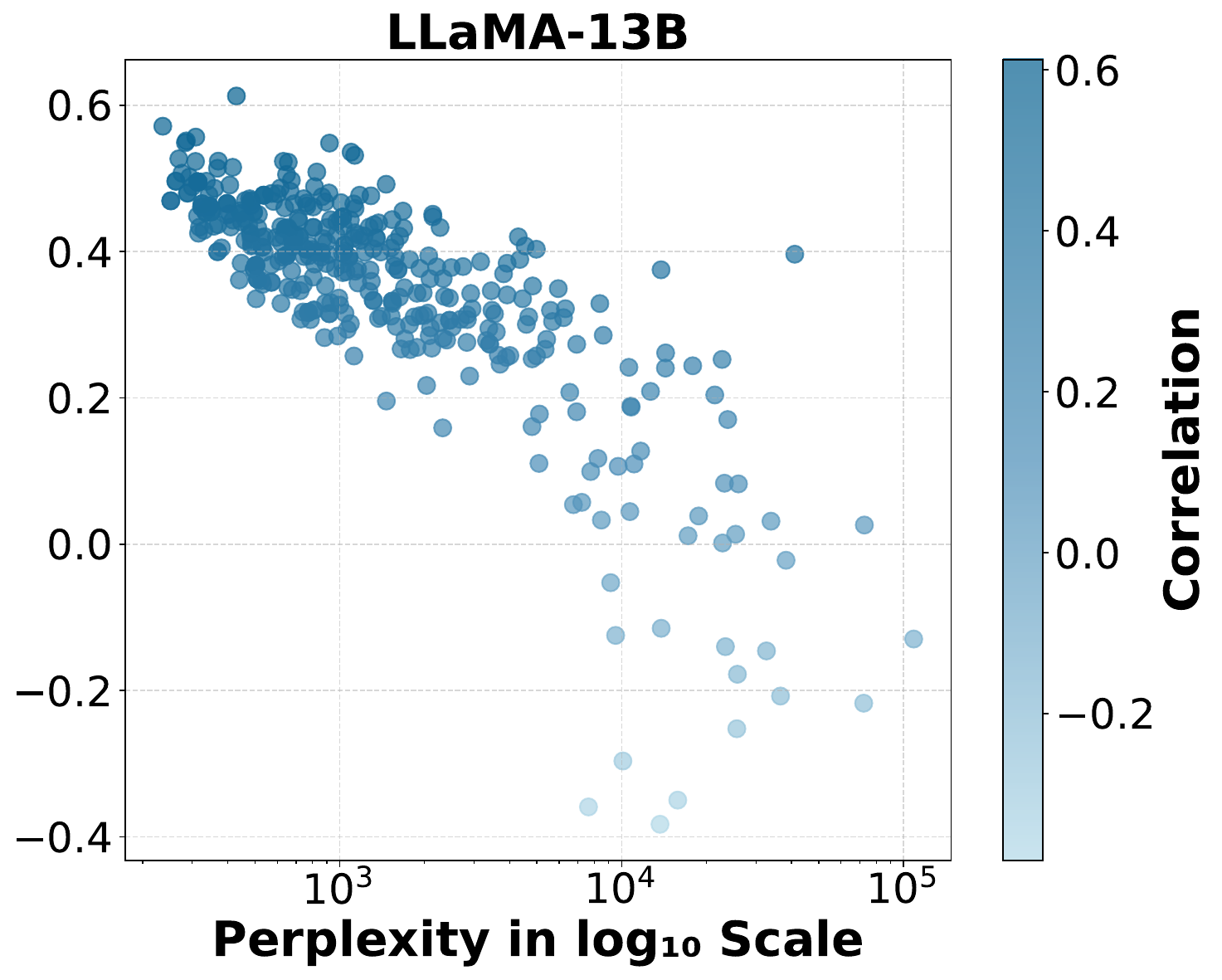}
    \caption{The correlation between sparsity levels and Hessian traces influence individuals' perplexity.}
    \label{fig:correlation}
\end{figure}

\noindent \textbf{Robustness to Calibration Sample Size.}
SparseGPT, Wanda and our MSP all require calibration data for obtaining either input activations or the Hessian matrix. To assess the robustness of our algorithm with respect to the calibration data, we conduct a sensitivity test involving variations in the number of calibration samples in \cref{fig:ablation_nsamples}. A distinct trend emerges as the calibration sample size changes: All pruning methods demonstrate increased robustness and lower perplexity. However, with our MSP, SOTA pruning methods demonstrate more robustness to calibration samples than original performance. Finally, we choose 128 as the calibration size to ensure alignment with Wanda \cite{sun2023simple} and SparseGPT \cite{frantar2023sparsegpt}.

\subsection{Analysis}

\noindent \textbf{Correlation Analysis.}
In \cref{fig:loss}, we impose the perturbations on layers of different traces of Hessian matrix. The results show that, under the same perturbation, layers with higher trace values are affected more than those with lower trace values, indicating that the sensitivity of layers can be characterized by the trace of the Hessian matrix. Therefore, layers with large Hessian traces should not be pruned too much, and the sparsity ratios across layers should align with their corresponding trace values. We compute the correlations between the traces and sparsity ratios generated by our MSP to analyze how they influence perplexity performance. 
In \cref{fig:correlation}, we plot the trace-sparsity correlation against the perplexity of pruned LLaMA-13B models. Our findings indicate that perplexity escalates substantially when the layer-wise sparsity ratios are suboptimally configured, particularly when highly sensitive layers undergo extensive pruning.

\noindent \textbf{Extreme Sparsity Analysis.}
In \cref{table:extremely_pruning_results}, we evaluate the extreme sparsity on LLaMA-13B and observe that  SOTA methods using our MSP consistently outperform original performance beyond 50\% sparsity. As the sparsity increases, the performance gap widens, which aligns with the observations in \cref{fig:loss}. At lower sparsity, perturbations are small, resulting in a small loss. However, at higher sparsity levels, perturbations become significant. When previous SOTA methods apply uniform sparsity across all layers, this can destroy critical layers. In contrast, integrating our MSP enhances the stability and performance of the original methods, particularly at extreme sparsity levels.

\section{Conclusion}
In this work, we propose an efficient evaluation method based on the trace of the FIM to quantitatively measure and verify the varying sensitivities across layers. We introduce MSP, a pruning-oriented evolutionary algorithm designed to determine the optimal layer-wise sparsity levels. To enhance convergence speed and improve performance, we incorporate an efficient sensitivity-informed initialization strategy for the EA population. Moreover, MSP serves as a plug-and-play module, enabling integration with existing pruning methods. Extensive experiments demonstrate that our approach significantly improves the performance of state-of-the-art pruning techniques across various sparsity levels,  in particular, at extreme pruning ratios.

\section{Limitations}
The existing pruning methods for LLM mainly focus on designing different metrics to measure the importance of network components. In this work, our MSP approaches pruning from a different perspective: optimizing the layer-wise sparsity. MSP can be integrated into many existing methods to improve their performance with optimal layer-wise sparsity. However,  our MSP uses evolutionary optimization to determine the layer-wise sparsity, introducing additional computational costs. Considering the significant performance improvements, in particular, on extreme sparsity ratios, we believe the additional computations are worthwhile. 



\bibliography{latex/main}
\bibliographystyle{unsrt}

\clearpage
\appendix
\section{Detailed Pruning-oriented Evolutionary Algorithm}\label{asec:detail_ea}

\subsection{Random Initialization}
\label{asubsec:random_initialization}

As outlined in \cref{alg:initialize_population}., the population consists of a group individuals, each representing a potential configuration of layer sparsity for the given model. Given a target pruning ratio of N:M, the pruning ratios are initialized such that M matches the target sparsity, while N is randomly generated. Specifically, each individual is a vector of pruning ratios, where each element corresponds to a layer in the model and its value indicates the pruning level of that layer. Despite the randomness in N, the overall sparsity constraint is preserved, ensuring that the total pruning ratio across all layers aligns with the target pruning ratio of N:M. The overall sparsity ratio is maintained at the target value by a straightforward algorithm. This initialization strategy provides diversity in the search space while maintaining the global sparsity requirement. This constraint ensures that the population starts with valid configurations that meet the desired sparsity for each layer of the model. 

\begin{algorithm}[!ht]
    \caption{Random Initialization}
    \label{alg:initialize_population}
    \begin{algorithmic}
    \STATE {\bfseries Input:} Target sparsity ratio $N:M$, Population size $P$, Layers of models $num\_layers$
    \STATE {\bfseries Output:} Initialized population
    \end{algorithmic}
    \begin{algorithmic}[1]
    \FOR{$i = 1$ to $P$}
      \WHILE{True}
        \FOR{$j=1$ {\bfseries to} $num\_layers$} 
            \STATE $ind\gets rand(0,M)$
        \ENDFOR
         \IF{$\sum(ind) == N \times num\_layers$}
            \STATE append $ind$ to $popu$
            \STATE \textbf{break}
         \ENDIF
      \ENDWHILE
   \ENDFOR
   \STATE \textbf{return} $popu$
\end{algorithmic}
\end{algorithm}

\subsection{Sensitivity-informed Initialization}
\label{asubsec:sensitivity-inspired_initialization}

\begin{algorithm}[!ht]
   \caption{Sensitivity-informed Initialization}
   \label{alg:sensitivity_init}

    \begin{algorithmic}
   \STATE \textbf{Input:} $N, M$, $\text{num\_layers}, \text{population\_size}$
   \STATE \textbf{Output:} Initialized population
   \end{algorithmic}
   \begin{algorithmic}[1]
   \STATE Initialize $\Delta_{\text{front}}$ , $\Delta_{\text{deeper}}$
   \STATE $\text{num\_front\_layers} \gets \lfloor \text{num\_layers} / 5 \rfloor$
   \FOR{$i = 1$ to $\text{population\_size}$} 
        \WHILE{True}
           \STATE Sample $\delta_{\text{front}}$ from $\{0,1\}^{\text{num\_front\_layers}}$
           \STATE Compute $S_{\text{front}} \gets \sum \delta_{\text{front}}$
           \IF{$S_{\text{front}} \leq(\text{num\_deeper\_layers}) (M - N)$}
           \STATE Append $\delta_{\text{front}}$ to $\Delta_{\text{front}}$
           \STATE \textbf{break}
           \ENDIF
        \ENDWHILE
   \ENDFOR
   \FOR{$i = 1$ to $\text{population\_size}$} 
        \WHILE{True}
           \STATE Sample $\delta_{\text{deeper}}$
           \STATE Compute $S_{\text{deeper}} \gets \sum \delta_{\text{deeper}}$
            \IF{$S_{\text{deeper}} = S_{\text{front}}[i]$}
                \STATE Append $\delta_{\text{deeper}}$ to $\Delta_{\text{deeper}}$
                \STATE \textbf{break}
            \ENDIF
       \ENDWHILE
   \ENDFOR
   \STATE Initialize $\text{population}$
   \FOR{$(\delta_{\text{front}}, \delta_{\text{deeper}})$ in $(\Delta_{\text{front}}, \Delta_{\text{deeper}})$}
       \STATE Compute $\delta \gets [-\delta_{\text{front}}, \delta_{\text{deeper}}]$
       \STATE Compute ${ind} \gets N  + \delta$
       \STATE Append ${ind}$ to $\text{population}$
   \ENDFOR
   \RETURN $\text{population}$
\end{algorithmic}
\end{algorithm}

In \cref{alg:sensitivity_init}, we split the $\Delta individual$ into two parts: the front $\frac{1}{5}$ and the deeper $\frac{4}{5}$. 
The front part, $\Delta front$, is initialized as:
\begin{equation}
    \Delta_{\text{front}} = [\delta_1, \delta_2, \dots, \delta_k], \quad \delta_i \in \{0,1\},
\end{equation}
with the constraint:
\begin{equation}
    \sum_{i=1}^{k} \delta_i \leq (L - k) (M - N),
\end{equation}
where $L$ is the number of layers, and $k$ represents the $k$ layers in the front part. In our sensitivity-informed initialization, $k$ is set to $\frac{1}{5}$ of $L$. The deeper part, $\Delta deeper$, is initialized as:
\begin{equation}
\begin{split}
    \Delta_{\text{deeper}} =
    [\delta_{k+1}, \delta_{k+2}, \dots, \delta_L],
    \\\quad \delta_j \in \{0,1,\dots,M-N-1\}.
\end{split}
\end{equation}
 with the constraint that ensures the average sparsity is equal to the target sparsity:
\begin{equation}
    \sum_{j=k+1}^{L} \delta_j = \sum_{i=1}^{k} \delta_i.
\end{equation}
Finally, we concatenate $\Delta front$ and $\Delta deeper$ as:
\begin{equation}
    \Delta = [-\Delta_{\text{front}}, \Delta_{\text{deeper}}],
\end{equation}
The individual is then defined as:
\begin{equation}
    \text{individual} = NL + \Delta.
\end{equation}

\subsection{Crossover}
\label{asubsec:crossover}
In \cref{alg:crossover}, crossover is performed between two parent individuals selected from the population. This process involves randomly selecting a crossover point in the individual and swapping the subsequences beyond this point between two parents to create two offspring. The crossover operation ensures that the offspring inherit characteristics from both parents, promoting genetic diversity within the population. 

\begin{algorithm}[!ht]
   \caption{Crossover}
   \label{alg:crossover}
\begin{algorithmic}
   \STATE {\bfseries Input:} Parent $p1$, $p2$, Target sparsity $N:M$, Number of layers $num\_layers$
   \STATE {\bfseries Output:} Two offspring $child1$, $child2$
   \end{algorithmic}
   \begin{algorithmic}[1]
   \STATE randomly select $point \in [1, \text{len}(parent1) - 1]$
   \STATE $child1 \gets p1[:point] + p2 [point:]$
   \STATE $child2 \gets p2[:point] + p1[point:]$
   \FOR{ each  $child \in [child1, child2]$}
   \WHILE{True}
      \STATE Compute $average\_sparsity$
      \IF{$avg\_sparsity == N:M$}
         \STATE \textbf{break}
      \ELSIF{$avg\_sparsity < N:M$}
         \STATE Select an $index$
         \IF{$child[index] < M$}
            \STATE $child[index] \gets child1[index] + 1$
         \ENDIF
      \ELSIF{$avg\_sparsity > N:M$}
         \STATE Select an $index$
         \IF{$child[index] \geq 1$}
            \STATE $child[index] \gets child[index] - 1$
         \ENDIF
      \ENDIF
   \ENDWHILE
   \ENDFOR
   \STATE \textbf{return} $child1, child2$
\end{algorithmic}
\end{algorithm}

\subsection{Mutation}
\label{asubsec:mutation}
The function of mutation as shown in \cref{alg:mutation} introduces random variations to an individual with a probability defined by the mutation\_rate. Two sparsity ratios of random layers are selected, and their values are replaced with random sparsity ratios within the range from 0:M to M:M, like 0, 1:M, 2:M, ..., M:M.  To maintain the validity of the individual, the mutation is also constrained such that the average pruning ratio across all layers equals the target sparsity for each individual. The mutation process ensures that the algorithm can explore regions of the search space that might not be reached through crossover alone. This step helps to prevent invalid solutions while allowing for further exploration of the search space.

\begin{algorithm}[!ht]
   \caption{Mutation}
   \label{alg:mutation}
\begin{algorithmic}
   \STATE {\bfseries Input:} Individual $ind$, Target sparsity $N:M$
   \STATE {\bfseries Output:} Mutated $ind$
   \end{algorithmic}
   \begin{algorithmic}[1]
   \STATE Randomly select two indices $idx\_1$ and $idx\_2$ from $[0, num\_layers - 1]$
   \IF{$\text{random\_number} < \text{mutation\_rate}$}
       \WHILE{True}
           \STATE Save original values: 
           
           $original\_1 \gets ind[idx\_1]$, 
           
           $original\_2\gets ind[idx\_2]$
           \STATE Mutate: $ind[idx\_1] \gets \text{rand}(0, M)$
           \STATE Mutate: $ind[idx\_2] \gets \text{rand}(0, M)$
           \IF{$avg\_sparsity == N:M$}
               \STATE \textbf{Break}
           \ENDIF
           \STATE Revert: $ind[idx\_1] \gets original\_1$,
           
           $ind[idx\_2] \gets original\_2$
       \ENDWHILE
   \ENDIF
   \STATE \textbf{return} $ind$
\end{algorithmic}
\end{algorithm}
\section{Expanding the Experimental Results}
\label{asec:experimental_results}
\subsection{Perplexity of Previous LLM models}
\label{asubsec:opt_results}

\cref{table:msp_ppl_opt_r75} presents the perplexity of pruned OPT family models ranging from 125 million to 13 billion parameters at 75\% sparsity (M=4) on the WikiText dataset. After integrating with our MSP, all pruning methods achieve significantly improved performance compared to their original versions.

\begin{table*}[!ht]
\centering
\caption{WikiText perplexity (PPL) of pruned OPT family models with 75\% sparsity.}
\label{table:msp_ppl_opt_r75}
\resizebox{1\columnwidth}{!}{
\begin{tabular}{cccccccc}
\toprule[0.1em]
                         &                                  & \multicolumn{6}{c}{OPT}                                                                                         \\ \cline{3-8} 
\multirow{-2}{*}{Method} & \multirow{-2}{*}{Sparsity} & 125m             & 350m             & 1.3B             & 2.7B             & 6.7B             & 13B              \\ \hline
Dense                    & -                                & 27.65            & 22.00            & 14.62            & 12.47            & 10.86            & 10.13            \\ \hline
Magnitude                & 75\%(3:4)                        & 5585.35          & 11707.31         & 12056.72         & 17959.43         & 10513.13         & 175871.84        \\
\rowcolor[HTML]{EFEFEF} 
Magnitude + MSP          & 75\%(M=4)                        & \textbf{1515.40} & \textbf{2401.22} & \textbf{6350.61} & \textbf{8384.13} & \textbf{7710.47} & \textbf{8004.21} \\
SparseGPT                & 75\%(3:4)                        & 1383.65          & 1136.35          & 2570.75          & 475.57           & 1602.60          & 9840.13          \\
\rowcolor[HTML]{EFEFEF} 
SparseGPT + MSP          & 75\%(M=4)                        & \textbf{848.85}  & \textbf{538.65}  & \textbf{2012.85} & \textbf{300.52}  & \textbf{119.37}  & \textbf{167.55}  \\
Wanda                    & 75\%(3:4)                        & 2570.89          & 3100.23          & 2408.60          & 7300.13          & 2507.98          & 6036.00          \\
\rowcolor[HTML]{EFEFEF} 
Wanda + MSP              & 75\%(M=4)                        & \textbf{888.43}  & \textbf{831.88}  & \textbf{682.65}  & \textbf{552.78}  & \textbf{309.63}  & \textbf{301.99}  \\ \bottomrule[0.1em]
\end{tabular}
}
\end{table*}

\begin{table*}[!ht]
\centering
\caption{WikiText perplexity (PPL) of pruned LLMs with 50\% sparsity.}
\label{table:msp_ppl_main_result_r50}
\resizebox{0.8\columnwidth}{!}{
\begin{tabular}{ccccccccccccccc}
\toprule[0.1em]
                         &                            & \multicolumn{4}{c}{LLaMA}                                       & \multicolumn{3}{c}{LLaMA-2}                    \\ \cline{3-9} 
\multirow{-2}{*}{Method} & \multirow{-2}{*}{Sparsity} & 7B             & 13B            & 30B           & 65B           & 7B             & 13B           & 70B           \\ \hline
Dense                    & -                          & 5.68           & 5.09           & 4.10          & 3.53          & 5.12           & 4.57          & 3.12          \\ \hline
Magnitude                & 50\%(2:4)                  & 42.55          & 18.32          & 9.11          & \textbf{6.77} & 54.38          & 10.74         & \textbf{6.33} \\
\rowcolor[HTML]{EFEFEF} 
Magnitude + MSP          & 50\%(M=4)                  & \textbf{26.61} & \textbf{15.66} & \textbf{8.08} & 7.12          & \textbf{17.33} & \textbf{7.72} & 6.42          \\
SparseGPT                & 50\%(2:4)                  & 10.81          & 9.01           & 7.20          & 6.22          & 10.23          & 8.29          & 5.40          \\
\rowcolor[HTML]{EFEFEF} 
SparseGPT + MSP          & 50\%(M=4)                  & \textbf{10.38} & \textbf{8.05}  & \textbf{6.98} & \textbf{6.13} & \textbf{9.15}  & \textbf{7.46} & \textbf{5.35} \\
Wanda                    & 50\%(2:4)                  & 11.53          & 9.59           & 6.90          & 6.24          & 11.35          & 8.37          & 5.20          \\
\rowcolor[HTML]{EFEFEF} 
Wanda + MSP              & 50\%(M=4)                  & \textbf{10.29} & \textbf{8.02}  & \textbf{6.71} & \textbf{6.13} & \textbf{9.49}  & \textbf{7.37} & \textbf{5.19} \\ \hline
Magnitude                & 50\%(4:8)                  & 16.83          & 13.72          & 7.62          & \textbf{6.02} & 16.53          & 9.24          & 5.54          \\
\rowcolor[HTML]{EFEFEF} 
Magnitude + MSP          & 50\%(M=8)                  & \textbf{14.97} & \textbf{11.97} & \textbf{7.21} & 6.20          & \textbf{11.78} & \textbf{6.61} & \textbf{5.37} \\
SparseGPT                & 50\%(4:8)                  & 8.52           & 7.38           & 6.14          & \textbf{5.31} & 7.99           & 6.58          & \textbf{4.59} \\
\rowcolor[HTML]{EFEFEF} 
SparseGPT + MSP          & 50\%(M=8)                  & \textbf{8.35}  & \textbf{7.12}  & \textbf{6.05} & 5.35          & \textbf{7.60}  & \textbf{6.49} & 4.63          \\
Wanda                    & 50\%(4:8)                  & 8.56           & 7.40           & 5.98          & 5.30          & 8.08           & 6.55          & 4.50          \\
\rowcolor[HTML]{EFEFEF} 
Wanda + MSP              & 50\%(M=8)                  & \textbf{8.20}  & \textbf{6.96}  & \textbf{5.86} & \textbf{5.29} & \textbf{7.52}  & \textbf{6.32} & \textbf{4.49} \\ \hline
Magnitude                & 50\%(8:16)                 & 14.58          & 12.65          & 7.15          & 5.69          & 13.74          & 8.46          & 5.28          \\
\rowcolor[HTML]{EFEFEF} 
Magnitude + MSP          & 50\%(M=16)                 & \textbf{13.74} & \textbf{11.93} & \textbf{6.80} & \textbf{5.69} & \textbf{12.40} & \textbf{6.23} & \textbf{5.15} \\
SparseGPT                & 50\%(8:16)                 & 7.80           & 6.75           & 5.69          & 4.99          & 7.11           & 6.05          & 4.30          \\
\rowcolor[HTML]{EFEFEF} 
SparseGPT + MSP          & 50\%(M=16)                 & \textbf{7.79}  & \textbf{6.66}  & \textbf{5.63} & \textbf{4.95} & \textbf{6.93}  & \textbf{5.97} & \textbf{4.28} \\
Wanda                    & 50\%(8:16)                 & 7.85           & 6.73           & 5.60          & 4.97          & 7.14           & 6.02          & 4.24          \\
\rowcolor[HTML]{EFEFEF} 
Wanda + MSP              & 50\%(M=16)                 & \textbf{7.63}  & \textbf{6.49}  & \textbf{5.48} & \textbf{4.90} & \textbf{6.87}  & \textbf{5.90} & \textbf{4.22} \\ \bottomrule[0.1em]
\end{tabular}
}
\vspace{-4mm}
\end{table*}

\subsection{Perplexity of Pruned LLMs at Lower Sparsity Ratio}
\label{asubsec:msp_lower_sparsity}

\cref{table:msp_ppl_main_result_r50} represents the perplexity of pruned LLaMA and LLaMA-2 models under 50\% sparsity. As shown in \cref{table:msp_ppl_main_result_r50}, When integrated with our MSP, these methods outperform their original performance on LLaMA-13B at a sparsity ratio of 50\%. However, their performance improvement is less pronounced compared to higher sparsity, such as 75\%. We hypothesize that this is because smaller perturbations have a relatively minor impact on the loss compared to larger ones (\cref{fig:loss}). Therefore, even when the optimal arrangement of sparsity levels across layers is identified, the performance improvement is not as pronounced as it is at higher sparsity.

\begin{figure}[!h]
    \centering
    \includegraphics[width=0.6\linewidth]{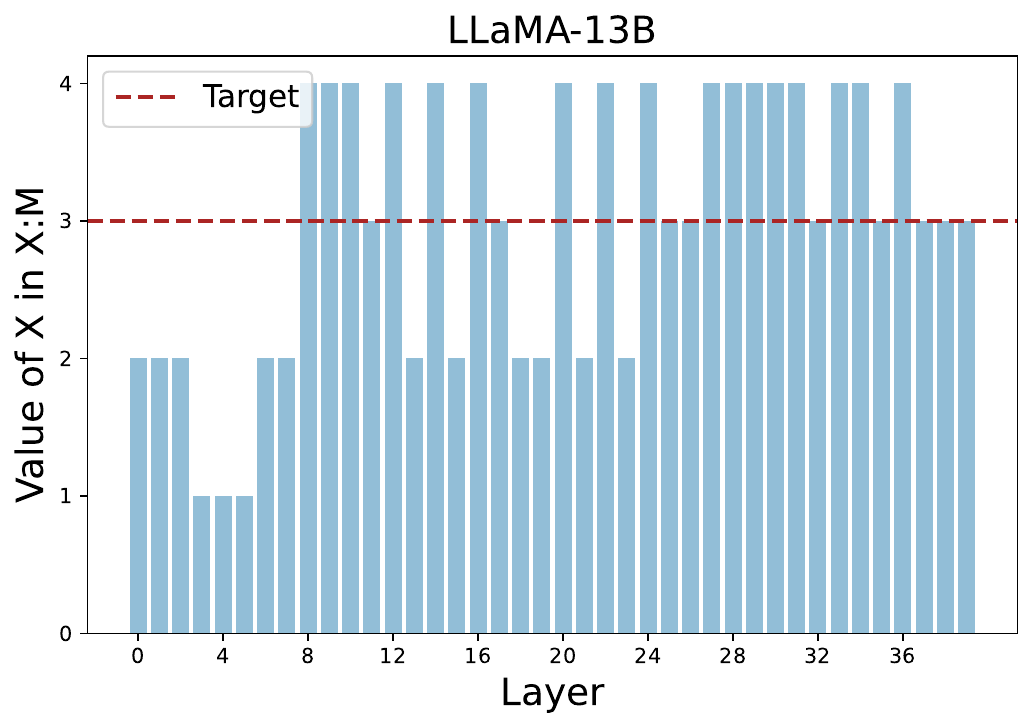}
    \caption{The sparsity levels of the optimal individual on pruned LLaMA-13B when target sparsity (N:M) is 3:4.}
    \label{fig:distribution}
    \vspace{-4pt}
\end{figure}

\begin{table*}[!ht]
\centering
\vspace{4mm}
\caption{Mean zero-shot accuracies (\%) of pruned LLaMA and LLaMA-2 models with 50\% sparsity.}
\label{msp_tasks_acc_main_result_r50}
\resizebox{0.9\columnwidth}{!}{
\begin{tabular}{ccccccccccccccc}
\toprule[0.1em]
                         &                            & \multicolumn{4}{c}{LLaMA}                                         & \multicolumn{3}{c}{LLaMA-2}                      \\ \cline{3-9} 
\multirow{-2}{*}{Method} & \multirow{-2}{*}{Sparsity} & 7B             & 13B            & 30B            & 65B            & 7B             & 13B            & 70B            \\ \hline
Dense                    & -                          & 60.01          & 62.60          & 65.37          & 66.99          & 59.68          & 63.02          & 67.13          \\ \hline
Magnitude                & 50\%(2:4)                  & 44.79          & 48.02          & 53.17          & 61.25          & \textbf{47.47} & 51.14          & \textbf{59.96} \\
\rowcolor[HTML]{EFEFEF} 
Magnitude + MSP          & 50\%(M=4)                  & \textbf{45.93} & \textbf{49.92} & \textbf{55.38} & \textbf{62.62} & 46.93          & \textbf{54.21} & 58.71          \\
SparseGPT                & 50\%(2:4)                  & 49.71          & 53.08          & \textbf{59.67} & 62.98          & \textbf{51.42} & \textbf{56.14} & \textbf{63.98} \\
\rowcolor[HTML]{EFEFEF} 
SparseGPT + MSP          & 50\%(M=4)                  & \textbf{50.50} & \textbf{53.75} & 58.94          & \textbf{62.99} & 50.30          & 55.94          & 63.32          \\
Wanda                    & 50\%(2:4)                  & 48.33          & 52.12          & \textbf{59.22} & 62.16          & 48.83          & 52.99          & \textbf{63.41} \\
\rowcolor[HTML]{EFEFEF} 
Wanda + MSP              & 50\%(M=4)                  & \textbf{49.48} & \textbf{53.52} & 59.18          & \textbf{63.46} & \textbf{49.57} & \textbf{54.65} & 63.04          \\ \hline
Magnitude                & 50\%(4:8)                  & 46.03          & 50.49          & 53.60          & 62.13          & 50.64          & 54.25          & 60.29          \\
\rowcolor[HTML]{EFEFEF} 
Magnitude + MSP          & 50\%(M=8)                  & \textbf{47.27} & \textbf{52.07} & \textbf{58.01} & \textbf{62.74} & \textbf{51.12} & \textbf{55.10} & \textbf{60.48} \\
SparseGPT                & 50\%(4:8)                  & 52.62          & \textbf{55.78} & \textbf{60.35} & \textbf{65.18} & \textbf{53.91} & \textbf{58.84} & \textbf{65.97} \\
\rowcolor[HTML]{EFEFEF} 
SparseGPT + MSP          & 50\%(M=8)                  & \textbf{53.08} & 55.77          & 59.16          & 64.81          & 53.17          & 58.50          & 65.70          \\
Wanda                    & 50\%(4:8)                  & \textbf{51.91} & \textbf{55.25} & \textbf{60.17} & \textbf{64.65} & 52.61          & \textbf{58.51} & \textbf{65.53} \\
\rowcolor[HTML]{EFEFEF} 
Wanda + MSP              & 50\%(M=8)                  & 49.29          & 55.09          & 59.93          & 63.81          & \textbf{53.41} & 57.42          & 65.38          \\ \hline
Magnitude                & 50\%(8:16)                 & \textbf{49.66} & 51.22          & 54.90          & 62.51          & 53.34          & 55.86          & 61.37          \\
\rowcolor[HTML]{EFEFEF} 
Magnitude + MSP          & 50\%(M=16)                 & 49.17          & \textbf{52.15} & \textbf{58.03} & \textbf{62.88} & \textbf{53.52} & \textbf{56.14} & \textbf{61.79} \\
SparseGPT                & 50\%(8:16)                 & 53.98          & \textbf{57.70} & 61.15          & \textbf{65.25} & \textbf{55.21} & \textbf{59.89} & \textbf{66.93} \\
\rowcolor[HTML]{EFEFEF} 
SparseGPT + MSP          & 50\%(M=16)                 & \textbf{54.48} & 57.62          & \textbf{61.99} & 65.13          & 54.43          & 58.67          & 66.67          \\
Wanda                    & 50\%(8:16)                 & 54.26          & 57.48          & \textbf{62.27} & 64.68          & \textbf{54.37} & 58.66          & \textbf{66.56} \\
\rowcolor[HTML]{EFEFEF} 
Wanda + MSP              & 50\%(M=16)                 & \textbf{54.34} & \textbf{57.84} & 61.87          & \textbf{64.99} & 54.32          & \textbf{59.88} & 65.73          \\ \bottomrule[0.1em]
\end{tabular}
}
\end{table*}

\begin{table*}[!ht]
\centering
\caption{Optimal layer-wise pruning sparsity levels for LLaMA-13B and their corresponding perplexity under different target sparsity.}
\label{Optimal_Pruning_Magnitude_of_llama13b}
\resizebox{1\columnwidth}{!}{
\begin{tabular}{cccc}
\toprule[0.1em]
Model                                                                              & Sparsity     & Optimal Layer-Wise Pruning Sparsity Levels                                                                                                                                                           & PPL     \\ \hline
\multirow{10}{*}{\begin{tabular}[c]{@{}c@{}}LLaMA-13B\\ (dense 5.09)\end{tabular}} & 50\%(M=4)    & {[}2, 1, 2, 1, 1, 1, 2, 2, 2, 2, 2, 2, 2, 2, 2, 2, 2, 2, 2, 2, 2, 2, 2, 2, 3, 2, 3, 2, 2, 2, 2, 2, 3, 2, 2, 2, 3, 2, 2, 2{]}                                                                              & 8.02    \\
                                                                                   & 50\%(M=8)    & {[}4, 3, 3, 3, 3, 4, 4, 3, 4, 4, 4, 4, 4, 4, 4, 4, 5, 4, 4, 4, 5, 3, 5, 4, 4, 4, 4, 5, 4, 5, 4, 4, 5, 4, 4, 4, 4, 4, 4, 4{]}                                                                              & 6.96    \\
                                                                                   & 50\%(M=16)   & {[}7, 7, 7, 7, 7, 8, 8, 6, 8, 8, 8, 8, 8, 8, 8, 8, 9, 8, 8, 8, 9, 8, 8, 8, 8, 8, 9, 9, 9, 8, 10, 8, 8, 8, 7, 9, 8, 8, 8, 8{]}                                                                             & 6.49    \\
                                                                                   & 62.5\%(M=8)  & {[}3, 4, 3, 2, 5, 4, 4, 4, 6, 4, 5, 5, 5, 4, 5, 5, 5, 8, 4, 4, 5, 5, 7, 5, 8, 5, 5, 5, 5, 5, 5, 8, 5, 5, 5, 6, 5, 5, 6, 6{]}                                                                              & 12.21   \\
                                                                                   & 62.5\%(M=16) & \begin{tabular}[c]{@{}c@{}}{[}6, 8, 6, 4, 10, 8, 8, 8, 12, 8, 10, 10, 10, 8, 10, 10, 10, 16, 8, 8, 10, 10, \\ 14, 10, 16, 10, 10, 10, 10, 10, 10, 16, 10, 10, 10, 12, 10, 10, 12, 12{]}\end{tabular}      & 10.48   \\
                                                                                   & 75\%(M=4)    & {[}2, 2, 2, 1, 1, 1, 2, 2, 4, 4, 4, 3, 4, 2, 4, 2, 4, 3, 2, 2, 4, 2, 4, 2, 4, 3, 3, 4, 4, 4, 4, 4, 3, 4, 4, 3, 4, 3, 3, 3{]}                                                                              & 52.76   \\
                                                                                   & 75\%(M=8)    & {[}5, 4, 1, 4, 4, 3, 4, 4, 8, 8, 8, 8, 8, 4, 5, 6, 5, 8, 5, 4, 8, 4, 6, 5, 8, 8, 7, 8, 8, 5, 6, 8, 7, 8, 6, 8, 6, 6, 6, 6{]}                                                                              & 35.66   \\
                                                                                   & 75\%(M=16)   & \begin{tabular}[c]{@{}c@{}}{[}6, 10, 6, 6, 3, 11, 7, 6, 8, 13, 16, 16, 12, 7, 13, 16, 16, 12, 12, 10, 16, 10,\\  13, 16, 16, 16, 13, 13, 14, 16, 12, 16, 14, 12, 10, 16, 16, 11, 11, 13{]}\end{tabular}   & 34.51   \\
                                                                                   & 87.5\%(M=8)  & {[}4, 4, 4, 5, 5, 4, 6, 8, 6, 7, 8, 8, 8, 7, 8, 8, 8, 7, 8, 7, 8, 7, 7, 8, 8, 7, 8, 6, 7, 8, 7, 8, 8, 7, 8, 8, 7, 8, 8, 7{]}                                                                              & 1421.25 \\
                                                                                   & 87.5\%(M=16) & \begin{tabular}[c]{@{}c@{}}{[}8, 8, 8, 8, 8, 8, 10, 16, 14, 15, 16, 16, 16, 14, 16, 16, 16, 14, 16, 14, 16, 14,\\  14, 16, 16, 14, 16, 15, 14, 16, 14, 16, 16, 14, 16, 16, 14, 16, 16, 14{]}\end{tabular} & 1160.59 \\ \bottomrule[0.1em]
\end{tabular}
}
\vspace{-4mm}
\end{table*}

\subsection{Zero-shot Tasks Accuracy of Pruned LLMs at Lower Sparsity Ratio}
\label{asubsec:zero-shot_lower_sparsity}
In \cref{msp_tasks_acc_main_result_r50}, we show the mean zero-shot accuracies of pruning methods integrated with our MSP and baselines across seven zero-shot tasks on pruned LLaMA and LLaMA-2 models. Given that no fine-tuning takes place, there is an obvious gap between sparse pruned LLMs and the den  se LLMs. As the model size increases, however, the gap of accuracy diminishes. The average accuracy of LLaMA-2-7B pruned by Wanda \cite{sun2023simple} with our MSP integration reaches 49.57, significantly outperforming its original performance.

\subsection{Searched Optimal Layer-wise Pruning Sparsity Levels with Perplexity}
\label{asubsec:searched_sparsity}

\cref{Optimal_Pruning_Magnitude_of_llama13b} presents the optimal layer-wise pruning sparsity levels for LLaMA-13B, along with their corresponding perplexity under different target sparsity levels. Notably, all selected individuals exhibit lower sparsity than the target in the top $\frac{1}{5}$ of the layers. For example, as shown in \cref{fig:distribution}, layers 0 to 7 have sparsity levels lower than the target sparsity (3:4) represented by the red dotted line.

\end{document}